# Transformer Models in Education: Summarizing Science Textbooks with AraBART, MT5, AraT5, and mBART

Sari Masri[1,*], Yaqeen Raddad[1,*], Fidaa Khandaqji[1,*], Huthaifa I. Ashqar[2], Mohammed Elhenawy [3]

[*] Authors contributed equally to this work.

[1] Department of Natural, Engineering and Technology Sciences, Arab American University, Ramallah P.O. Box 240, Palestine

[2] Civil Engineering Department, Arab American University, Jenin P.O. Box 240, Palestine and Columbia University, NY, USA

`Huthaifa.ashqar@aaup.edu`

[3] CARRS-Q, Queensland University of Technology, Brisbane, Australia

**Abstract.** Recently, with the rapid development in the fields of technology and the increasing amount of text t available on the internet, it has become urgent to develop effective tools for processing and understanding texts in a way that summaries the content without losing the fundamental essence of the information. Given this challenge, we have developed an advanced text summarization system targeting Arabic textbooks. Relying on modern natural language processing models such as MT5, AraBART, AraT5, and mBART50, this system evaluates and extracts the most important sentences found in biology textbooks for the 11$^{th}$ and 12$^{th}$ grades in the Palestinian curriculum, which enables students and teachers to obtain accurate and useful summaries that help them easily understand the content. We utilized the Rouge metric to evaluate the performance of the trained models. Moreover, experts in education Edu textbook authoring assess the output of the trained models. This approach aims to identify the best solutions and clarify areas needing improvement. This research provides a solution for summarizing Arabic text. It enriches the field by offering results that can open new horizons for research and development in the technologies for understanding and generating the Arabic language. Additionally, it contributes to the field with Arabic texts through creating and compiling schoolbook texts and building a dataset.

## 1 Introduction

Text summarization technology is fundamental and has many uses such as information retrieval and language analysis. Summarization can be divided into two types: Abstractive and Extractive. Extractive summarization creates summaries using exact words from the original text, while abstractive summarization generates new sentences or paraphrases, often using new words not found in the source [1]. As the volume of textual information on the Internet grows, it becomes increasingly challenging for readers to search through and find valuable content. This challenge has led researchers to develop automated text summarization technologies that generate concise summaries capturing the key information of texts. Although this field is rich in research, Arabic text studies face significant challenges due to the language's complexity and limited NLP (Natural Language Processing) resources. The intricate word structure in Arabic adds to the Morphological Complexity, complicating text analysis. Script Issues further challenge text segmentation due to the cursive and variable letter shapes in Arabic writing. Additionally, the lack of Annotated Corpora restricts the availability of essential data for developing robust NLP models, particularly for complex tasks like text summarization that demand a thorough linguistic and contextual grasp.

The field of Natural Language Processing (NLP) endeavors to develop techniques capable of handling various language tasks like translation, summarization, and generation. These tasks all require extracting meaning from text. However, the current single-task training approach produces limitations, as it requires creating specific models for each task and domain, leading to inefficiencies. Recurrent Neural Networks (RNNs) have been widely used in NLP but face challenges with long sentences and parallelization due to their inherent structure. To address these limitations, the Transformer model was introduced. Unlike RNNs, Transformers rely only on attention mechanisms, eliminating issues like vanishing gradients and enabling easier parallelization. This advancement facilitates the training of more flexible networks for NLP tasks [2].

Pre-training language models has proven to be beneficial in enhancing the performance of various natural language processing (NLP) tasks; this includes tasks at the sentence level, such as natural language inference and paraphrasing, which focus on identifying relational ships between sentences through holistic analysis. It also covers token-level tasks like named entity recognition and question answering, which provide detailed outputs for each token [3].

In this paper, we developed and refined a comprehensive dataset based on the Palestinian curriculum, for 11th and 12th-grade Biology, focusing on enhancing understanding of advanced linguistic models such as mBART50 [4], AraBART [5], MT5 [6] and AraT5 [7] for educational content. The dataset is diverse and rich in covering complex scientific concepts, providing a strong foundation for training deep language models. By overcoming challenges related to the organization and linguistic standardization of educational content, we have contributed innovative solutions that enhance the models' ability to accurately understand and generate educational texts.

Our contribution can be summarized as follows:

1. Collection and revision of a specialized dataset: A dataset has been created from the biology textbooks for the 11$^{th}$ and 12$^{th}$ grades of the Palestinian curriculum, with focused on the unique addition provided by the paper to the research field, this selection of the textbook was based on its comprehensive and accurate content, and the texts were carefully collected to ensure diversity of the curriculum ,this contribution is characterized by a focused on providing reliable and useful scientific content for research.
2. Training MT5, AraBART, AraT5 and mBART50 models to enhancing understanding of educational content and efficient text generation and summarization using the specialized datasets collected from the biology textbooks for the 11$^{th}$ and 12$^{th}$ grades of the Palestinian curriculum, MT5, AraBART, AraT5 and mBART50 models were trained , the training process was designed to improve the models understanding of the educational content presented in the textbooks and to increase their ability to generate and summarize texts efficiently. During the training process, the training parameters and performance metrics were fine-tuned based on the unique characteristics of the educational content, this contributed to improving the quality and accuracy of the models in understanding and generating summaries for educational materials.

3. Manual evaluation by an expert: this method involves evaluating summaries generated by pre-trained models where model expert assigns a rating from 1 to 10 based on specific criteria related to the quality of the summary, its accuracy, and its capture and of the key elements from the original text. We employed this approach to assess the summaries generated by the MT5, AraBART, AraT5 and mBART50 models. An expert with domain knowledge in the subject matter was tasked with evaluating the generated summaries. The expert assigned a rating of 7.17erall to the summary based on criteria such as coherence, informativeness and relevance to the original text.

This research paper stood out for its application of advanced training methodology focused on enhancing the model's performance in understanding educational content and generating comprehensive and accurate summaries. This paper begins with a careful review of previous studies on language representation, especially Arabic. The second section highlights the achievements and reveals the research gaps that our research intends to fill. The third section details the methodology used to develop the models, including the data collection and processing steps and the strategies used to train the models. In particular, we highlight how MT5, AraBART, AraT5 and mBART50 models can be modified. Section IV presents the subsequent tasks and datasets used to evaluate the model and provides a framework for understanding its performance. In Section V, we discussed the experimental setup and achieved results. We finally conclude our work in Section VI, presenting our conclusions and suggestions for future research, and emphasizing the importance of continued development and improvement along this path.

## 2      Literature Review

Most of the related work in text summarization has utilized extractive and abstractive approaches, which involve identifying important sentences in the original document to obtain the final summary. Al-Maleh et al [[8] on the utilization of deep learning techniques, employing extractive and abstractive summarization methods to summaries Arabic texts. El-shishrawy et al [[9] rephrased sentences to align with the research context and they used machine learning for text summarization. Sobhi et al [10] followed an extractive approach in extracting Arabic texts by dividing the problem into two sub-problems, identifying extracting sentences using machine learning techniques. On the other hand, utilized deep learning in natural language processing for text summarization, employing sequence-to-sequence models by Bahdabau et al [[11], and Sutskever et al [[12] neural networks were employed for text summarization, and the LSTM model was used for translating phrases from English to Arabic. Rush et al [13] The researchers utilized an attention-based approach for creative sentence summarization by employing deep learning techniques. In this method, each word in the summary is generated based on the original sentences in the original text. Reda et al [14] The researchers employed both extractive and abstractive summarization approaches to summarize articles using Arabert, they introduced a sequence-based model that relies on machine transformer techniques and the pre-trained MT5 model. In addition, the researchers in this study introduced pre-trained transfer learning models, such as BERT, which can perform various tasks including text summarization, question answering, and language inference. These models leverage deep learning techniques and have shown significant effectiveness in handling diverse natural language processing tasks by Devlin et al [[15] Adding an important contribution to previous studies is the summarization of Arabic texts based on a general engineering approach for natural language understanding (NLU) and natural language generation (NLG). This involves utilizing pre-trained transfer models such as AraBERT, BERT, XLNet, and XLM. The goal is to extract the most important sentences from the original text, enhancing the process of producing and comprehending natural language through NLU and NLG techniques by Abu Nada et al [16] , Wisam et al [17], researchers also trained transfer models (BERT) for processing Arabic texts, and the results showed advanced performance in most natural language processing tasks. Kieuvongngam et al [[18] the researchers utilized modern models such as BERT and GPT-2 to summarize medical articles related to COVID-19. ALmarjeh et al [19]researchers compared the model architectures based on RNN and how they were utilized in summarizing Arabic texts, such as BERT, AraGPT-2, AraT5.kahla et al [20] researchers developed models for summarizing extractive texts in Arabicusing the AraSum dataset, which contains 50,000 articles they trained and improved BART and mt5 models, and the results showed that these models outperformed previous models in this field dataset.

All previous studies must improve the understanding and processing of natural language, specifically addressing Arabic challenges. All studies employed extractive and abstractive approaches in text summarization, with some combined both Methods. They used transformer-based models such as BERT, GPT and AraT5, trained on substantial data. Previous studies focused on utilizing word embedding techniques, such as Word2Vec, to understand Arabic language models. In Arabic text generation, RNN models were first used, but BERT models surpassed them in quality, readability, and linguistic sequence. All studies used evaluation tools, such as ROUGE and F1-SCORE, to assess the performance of models. Some studies also relied on manual evaluation with the assistance of language experts, translating to English. The studies provided focus on the development of an extractive Arabic text summarizer using a general-purpose architecture for Natural Language Generation (NLG) and Natural Language Understanding (NLU) models, such as AraBART, BERT, XLNet, XLM, etc. The goal is to summarize Arabic documents by evaluating and extracting the most important sentences. The researchers propose an extractive summarizer based on AraBART and compare its efficiency using the Rouge measure and human evaluation.

## 3 Materials and Methods

The main aim of this study is to present and evaluate a new method for summarization of Arabic school textbooks by modifying and extending existing methods to include the use of mBART50, AraBART, MT5 and AraT5 models while applying improvement to these models to improve the summarization process, this improvement aim compares the results with expert human summarization to demonstrate the effectiveness and improved achieved. Figure 1 presents the structured methodology employed in our study to ensure systematic handling and analysis of the data collected. This methodology is segmented into four principal stages: Data Collection, Data Preparation, Modeling, and Summarization and Evaluation.

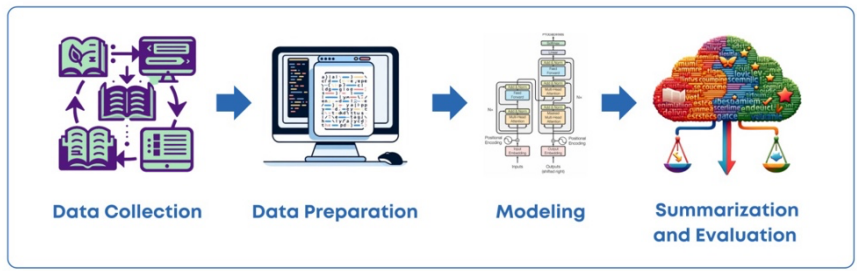

**Fig. 1.** Biology schoolbook summarization methodology.

### 3.1 Data Collection and Preparation

In the framework of our current project, which aims to develop a model for the synthesis of scientific texts, we have taken a strategic step by focusing on data collection from a rich and unique source: Biology school textbooks for $11^{th}$ and $12^{th}$ grades.

This approach to data collection is qualitative in the development of automated summary models. The text extraction process has been carried out with great care while maintaining the context and meaning of the original content. This process covered not only the basics but also graphics and interactive schemes that were possible and useful, the dataset we built consists of 6 features and 111 rows, the dataset scheme is as shown below.

**Table 1.** Table captions should be placed above the tables.

| Features | Description |
| --- | --- |
| id | A unique identifier for each section of the book |
| unit_title | The title of the educational unit within the book. |
| lesson_title | The title of the specific lesson within the unit. |
| section_content | The title of the section within the lesson. |
| questions | Questions related to the section content will enhance understanding. |
| expert_summary | The expert's summary to be used in the evaluation process. |

Data cleaning and preparation are crucial steps in enhancing the quality of the dataset and ensuring the effectiveness of training and accuracy of summarization for language models. In our research, a series of methodical techniques were applied to process the texts and summaries extracted from the dataset of school textbooks. Initially, we removed unnecessary marks that could distort the text. Secondly, we eliminated leading and trailing spaces and unified multiple spaces between words into a single space to achieve consistency in text formatting. Thirdly, to avoid complexities associated with varying case sensitivities, all English characters in texts and summaries were converted to lowercase. After these processes, any empty records were discarded to ensure incomplete data did not affect the quality and effectiveness of training. These methodical steps for data cleaning and preparation guarantee the creation of a unified and clean dataset, which contributes to enhancing the training capability and overall performance of the language model in the task of text summarization.

The figure below indicates that within the dataset, a significant number of entries have a word count exceeding 1000 in the section content feature. This results in a maximum input tokenizer length of 1024 for the section content having most of the observations into consideration, and 256 for the expert summary. The TokenizerFast library from Hugging Face is employed for tokenizing sentences. The dataset was divided into 70% for training, with the remaining 30% equally split between validation and testing.

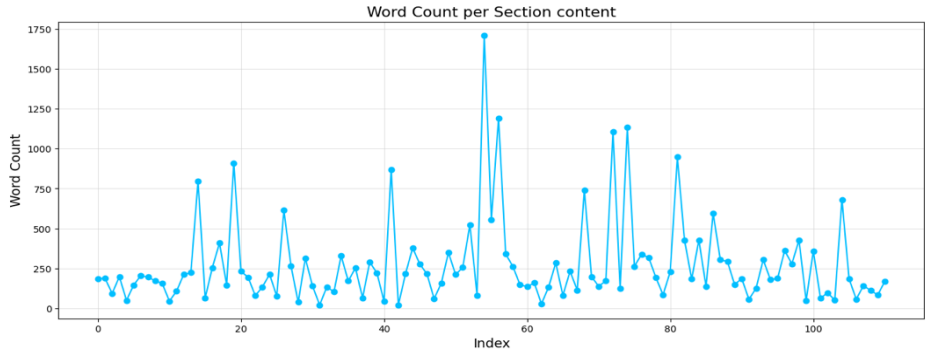

**Fig. 2.** The word count in section_content feature.

### 3.2 Modeling

After optimizing the mBART50, AraBART, MT5 and AraT5 models, they were directly used to generate summaries, this approach benefits from the advantages in text understanding and generation capabilities of the models to create concise and useful summaries, providing a valuable tool for students and teachers for quick access to essential information.

**Table 2.** Hyperparameters of the transformer model

| Hyperparameter | Value |
|---|---|
| Epoch | 10 |
| Learning rate | $5 \times 10^{-5}$. |
| Weight decay | 1% |
| Batch size | 4 |
| Optimizer | adamW |

In our research, we focused on exploring a range of advanced neural network models for Arabic text processing tasks, including translation and summarization. Each model was trained under a consistent set of hyperparameters: we employed a learning rate of $5 \times 10^{-5}$, conducted training over 10 epochs, and set a weight decay of 1%. The models were trained with a batch size of 4 using the AdamW optimizer, aiming to optimize performance while maintaining computational efficiency.

**mBART50**

mBART50, a multilingual sequence-to-sequence model, was utilized for Arabic text translation. Pre-trained in a diverse corpus covering 50 languages, mBART50 is particularly adept at understanding and translating contextual nuances in Arabic. We evaluated its performance by assessing the accuracy and fluency of the translations, ensuring they retained the semantic integrity of the original texts [4].

**AraBART**

AraBART, tailored specifically for the Arabic language, extends the BART architecture to support generative summarization tasks. Its unique adaptation to Arabic morphology and script nuances enhances its capability to produce coherent and concise summaries of Arabic text. The effectiveness of AraBART was measured using standard metrics like ROUGE scores, which helped in evaluating the relevance and information retention in the summaries [5].

**AraT5**

The Summarization-arabic-english-news model specializes in bilingual summarization, converting Arabic news articles into English summaries. This model bridges language barriers and facilitates understanding for non-Arabic speakers. We focused on the quality of the English summaries, evaluating them based on accuracy, readability, and the preservation of essential information [21].

**MT5**

The multilingual T5 (MT5) model was instrumental in summarizing Arabic texts, showcasing its versatility across multiple languages. We focused on MT5's summarization capabilities, evaluating the quality and coherence of its outputs. The performance assessments were based on ROUGE scores, highlighting the model's efficiency in producing relevant and high-quality summaries [6].

### 3.3 Evaluation

1. Manual evaluation by experts: This method relies on evaluating summaries generated by pre-training models, where the expert assigns a rating from 1 to 10 based on specific criteria related to the quality of the summary, its accuracy and its capture of the key elements from the original text.
2. Comparing expert summaries with pre-training models summaries and the original text: this method involves comparing the summaries prepared by experts with those generated by pre-training models, as well as considering the original text.
3. Using automated evaluation metrics:

$$ROUGE - 1 = \frac{\sum_{unigrams\,in\,reference} Count_{match}(unigram)}{\sum_{unigrams\,in\,reference} Count(unigram)}$$

$$ROUGE - 2 = \frac{\sum_{bigrams\,in\,reference} Count_{match}(bigram)}{\sum_{bigrams\,in\,reference} Count(bigram)}$$

$$ROUGE - L = \frac{Length\,of\,Longest\,Common\,Subsequence}{Length\,of\,Reference\,Summary}$$

ROUGE-1 assesses the degree of overlap between unigrams present in the generated summary and those found in the reference summaries. It quantifies the number of individual words from the generated summary that appear in the reference summary, thereby providing a straightforward metric for content overlap at the word level. ROUGE-2 assesses the degree of overlap between bigrams present in the generated summary and those found in the reference summaries. ROUGE-2 scores consecutive word pairs, providing insight into the coherence and ordering of information in the summaries and reflecting the extent to which phrases and more subtle content structures are captured by the summary system. ROUGE-L focuses on the longest common subsequence between the generated summary and the reference summaries. Unlike ROUGE-2, it does not require consecutive matches, but instead scores the longest sequence of words that occur in the same order in both texts. This measure is particularly useful for assessing the structural similarity of the summary to the reference. It provides insight into the extent to which the key thematic elements are retained, even if the exact wording differs.

## 4 Results and discussion

Table 3. Summarization Model Evaluation Metrics (Rouge score)

| Model | rouge1 | rouge2 | rougeL |
| --- | --- | --- | --- |
| AraBART | 0.2462 | 0.1184 | 0.2462 |
| mBART50 | 0.2431 | 0.1392 | 0.2431 |
| AraT5 (summarization-arabic-english-news) | 0.2065 | 0.1153 | 0.2065 |
| mt5 | 0.0566 | 0 | 0.0566 |

**Table 4.** Summarization Model Evaluation by Expert

| Model | Rating (0 – 10) |
|---|---|
| AraBART | 8.409 |
| mBART50 | 8.18 |
| AraT5 (summarization-arabic-english-news) | 7.727 |
| MT5 | 7.17 |

Table 3 provides a comprehensive overview of the performance of the different summary models evaluated using the ROUGE metric, which measures the overlap of n-grams between the generated summaries and the reference summaries. The metrics included in this analysis are ROUGE-1 (unigram overlap), ROUGE-2 (bigram overlap), ROUGE-L (longest common subsequence), and ROUGE-Lsum (summary-specific variant of ROUGE-L).

The models' performance as reflected in table 4 likely indicates their effectiveness in handling specialized academic content. AraBART, with the highest rating of 8.409, demonstrates a strong capability in accurately summarizing complex biological information, which is critical for educational tools and resources. The slightly lower scores for the other models, including mBART50 and AraT5, suggest that while they are also competent, they might not manage specific subtleties or specialized terms as precisely as AraBART. The ratings also imply that these models, including the mt5 with the lowest score of 7.17, could be useful in educational settings, aiding in the creation of concise study materials and summaries of dense textbook content.

The AraBART algorithm demonstrates robust performance across all ROUGE metrics, with scores of 0.2462 for ROUGE-1, ROUGE-L, and ROUGE-Lsum and a ROUGE-2 score of 0.1184. This indicates a notable degree of unigram and bigram overlap with the reference summaries. The degree of overlap with the reference summaries indicates that AraBART is capable of effectively capturing both the surface form and some finer details of the source texts. mBART50 also performs well, particularly in capturing bigram overlap, as evidenced by its ROUGE-2 score of 0.1392, the highest among the models tested. This indicates that mBART50 is particularly adept at comprehending and replicating intricate linguistic structures. Nevertheless, its overall performance in terms of unigrams and longest common subsequence is slightly inferior to that of AraBART.

mBART50 also performs well, particularly in capturing bigram overlap, as evidenced by its ROUGE-2 score of 0.1392, the highest among the models tested. This indicates that mBART50 is particularly proficient in comprehending and replicating intricate textual structures, although its overall performance in terms of unigrams and longest common subsequence is slightly inferior to that of AraBART.

The AraT5 model exhibits moderate effectiveness, with scores of 0.2065 for ROUGE-1, ROUGE-L, and ROUGE-Lsum and a ROUGE-2 score of 0.1153. Although the performance is satisfactory, it is not as robust as that of AraBART or mBART50, indicating that while essential content can be captured, there is room for improvement in capturing detailed content.

MT5 model shows limited performance in these evaluations, with low scores in all metrics. The scores for ROUGE-1 and ROUGE-L are only 0.06 and for ROUGE-2 0, meaning that there is no bigram overlap. This indicates that these models are not able to summarize the given Arabic texts in a way that matches well with the reference summaries. Their poor performance could be due to several factors, such as insufficient training data, suboptimal model tuning, or inherent limitations in handling the complex Arabic language.

**Table 5.** Comparison between summary of expert and models summary

| Model | Summary |
|---|---|
| Original text | قبيلة الإسفنجيات (المساميات) تعد الإسفنجيات من أبسط الكائنات التي تنتمي إلى المملكة الحيوانية، وهي في مقدّمة سُلّم تصنيف المملكة. نستخدم الإسفنج الصناعي في حياتنا اليومية للاغتسال، وتنظيف الأواني، وأسطح المكاتب، تُعد الإسفنجيات من الكائنات الحية متعدّدة الخلايا، وتُسمّى أيضاً المساميات؛ نظراً لكثرة المسامات في جسمها. تركيب الإسفنجيات: تتركب الإسفنجيات من طبقتين يفصل بينهما هلام متوسط, الطبقة الخارجيّة أشبه ما تكون بالنسيج الطلائي الذي يغطي الجسم، أما الطبقة الداخلية (السطح الداخلي) فهي تتكوّن من خلايا مطوّقة (سوطية) لها عدة وظائف، بينما يحتوي الهلام المتوسّط على مجموعة من الخلايا الأميبية. ويتخلل الطبقات عدداً كبيراً من الفتحات تُسمّى ثقوباً شهيقية، وفتحة زفيرية واحدة أو أكثر موجودة في أعلى الكائن. تكاثر الإسفنجيات: تتكاثر الإسفنجيات لا جنسيا وجنسيا. أولاً: التكاثر اللاجنسي 1- التجزؤ (التشظي): يحدث عند تعرض الإسفنج لقوة مؤثرة كبيرة، مثل مرور تيارات مائية؛ ما يؤدي إلى انفصال جزء منها ما يلبث أن يكوّن كائناً جديداً في منطقة أخرى. 2- التبرعم: ويحدث في جميع الظروف. 3- البريعمات (الدرائر): وتلجأ إليها إسفنجيات المياه العذبة خلال الظروف غير المناسبة، مثل الجفاف، أو تجمد المياه. ثانياً: التكاثر الجنسي: تتكاثر الإسفنجيات جنسياً من خلال إنتاج حيوانات منوية وبويضات من الخلايا الأميبية، وتنتقل الحيوانات المنوية من إسفنج إلى إسفنج آخر، حيث يتم الإخصاب داخلياً في الهلام المتوسط |
| Expert | قبيلة الاسفنجيات (المساميات): وهي من أبسط الكائنات الحية كالإسفنج نظرا لكثرة المسامات في جسمها. تتركب الإسفنجيات من طبقتين يفصل بينهما هلام متوسط, الطبقة الخارجيّة أشبه ما تكون بالنسيج الطلائي الذي يغطي الجسم، أما الطبقة الداخلية فهي تتكوّن من خلايا سوطية، بينما يحتوي الهلام المتوسّط على مجموعة من الخلايا الأميبية. وتتكاثر الإسفنجيات لا جنسيا من خلال التجزؤ أو التبرعم أو البريعمات, وجنسيا من خلال إنتاج حيوانات منوية وبويضات من الخلايا الأميبية |
| MT5 | قبيلة الإسفنجيات (المساميات): وهي من أبسط الكائنات الحية متعدّدة الخلايا، وتُسمّى أيضاً المساميات؛ نظراً لكثرة المسامات في جسمها. تصنيف المملكة الحيوانية: وهي في مقدّمة سُلّم تصنيفها الحيواني. تركيب الإسفنجيات: تتركب الإسفنجيات من طبقتين يفصل بينهما هلام متوسط, الطبقة الخارجيّة أشبه ما تكون بالنسيج الطلائي الذي يغطي الجسم، أما الطبقة الداخلية فهي تتكوّن من خلايا مطوّقة (سوطية) لها عدة وظائف, بينما يحتوي الهلام المتوسّط على مجموعة من الخلايا الأميبية |
| AraBART | قبيلة الإسفنجيات (المساميات): وهي كائنات حية متعددة الخلايا, وتسمى أيضا المساميات؛ نظرا لكثرة المسامات في جسمها. مكوناتها: تتركب من طبقتين يفصل بينهما هلام متوسط, الطبقة الخارجية أشبه ما تكون بالنسيج الطلائي الذي يغطي الجسم, أما الطبقة الداخلية (السطح الداخلي) فهي تتكون من خلايا مطوقة (سوطية) لها عدة وظائف بينما يحتوي الهلام المتوسط على مجموعة من الخلايا الأميبية. ويتخلل الطبقات عددا كبيرا من الفتحات تسمى ثقوبا شهيقية، وفتحة زفيرية واحدة أو أكثر موجودة في أعلى الكائن. ثانيا: التكاثر اللاجنسي: تتكاثر الإسفنج لا جنسيا وجنسيا من خلال إنتاج حيوانات منوية وبويضات من الخلايا الاميبية وتنتقل الحيوانات المنوية من إسفنج إلى إسفنج آخر حيث يتم الإخصاب داخليا في الهلام المتوسط |
| AraT5 (summarization-arabic-english-news) | تعد الإسفنجيات من أبسط الكائنات التي تنتمي إلى المملكة الحيوانية، وهي في مقدّمة سُلّم تصنيف المملكة. تتركب الإسفنجيات من طبقتين يفصل بينهما هلام متوسط, الطبقة الخارجيّة أشبه ما تكون بالنسيج الطلائي الذي يغطي الجسم، أما الطبقة الداخلية فهي تتكوّن من خلايا مطوّقة (سوطية) لها عدة وظائف، بينما يحتوي الهلام المتوسّط على مجموعة من الخلايا الأميبية |

The analysis of the model-generated summaries presented in Table 5 provides a comprehensive assessment of the capabilities of different AI models in synthesizing specialized biological content related to sponges. Each model displays specific strengths and demonstrates varying abilities in addressing complex scientific summarization tasks. The expert summary serves as a standard for the efficient summarization of key anatomical and reproductive information about sponges. This summary serves as a standard against which the accuracy and conciseness of the summaries produced by the AI can be assessed. The MT5 model provides a comprehensive summary of the anatomical structure and taxonomic classification of sponges. The approach of this model ensures thorough coverage of the relevant details, although it could benefit from refinement to improve conciseness and maintain the focus observed in the expert summary.

The AraBART algorithm performs commendably, producing a summary that is both comprehensive and closely aligned with the expert's summary. The summary deftly incorporates essential structural and reproductive details, and the inclusion of specific anatomical features such as inhalation and exhalation openings adds depth to the summary, even if it goes beyond the core elements prioritized in the expert summary. The summary provided by Summarization-Arabic-English-News is coherent and accurately reflects the anatomical structure and classification within the animal kingdom. While the summary effectively captures primary structure, expanding its scope to address reproductive strategies more comprehensively could bring it more closely in line with the expert summary and thus improve its completeness.

In conclusion, the performance of these models demonstrates the potential and existing capabilities of AI in the automated summarization of specialized scientific content. Models such as AraBART and mt5 exhibit robust summarization capabilities, suggesting that minor adjustments could enhance their alignment with expert-generated summaries. The summarization models, and namely Summarization-Arabic-English-News, demonstrate potential for improvement, particularly in ensuring comprehensive coverage of all critical aspects of the content. This evaluation highlights the necessity of continuous refinement and domain-specific training to enhance the precision and relevance of AI-generated summaries in specialized fields such as biology. These findings contribute to the ongoing discourse on the further development and optimization of AI-driven summarization tools in the academic and research environment.

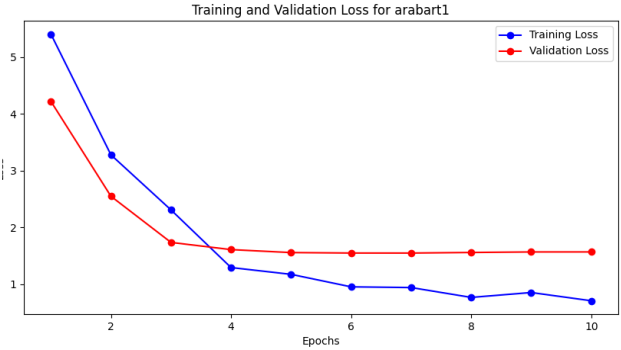

**Fig. 3.** Training and validation loss for arabert.

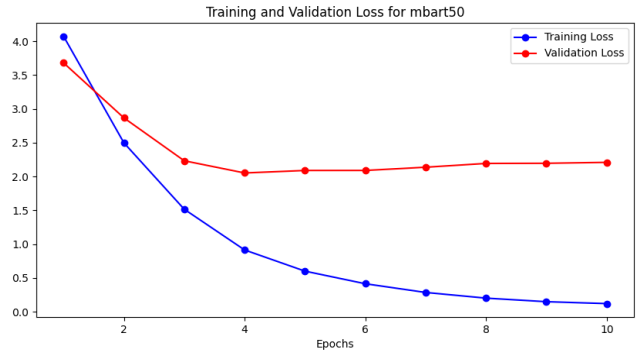

**Fig. 4.** Training and validation loss for mBart50.

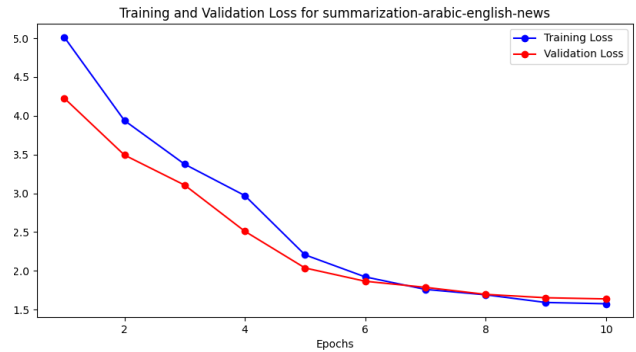

**Fig. 5.** Training and validation loss for summarization arabic news model.

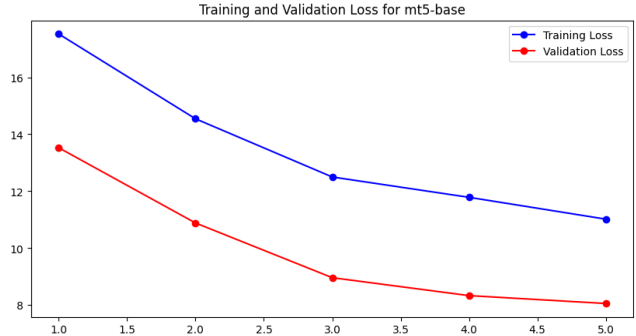

**Fig. 6.** Training and validation loss for mt5 model.

In Figures 3, 4, 5, and 6, the training and validation loss graphs of the arabert, mbart50, AraT5 (summarization-arabic-english-news), and mt5 models reveal varied learning behaviors and model efficiencies. Figure 3 (arabert) and Figure 5 (summarization-arabic-english-news) show promising trends where both training and validation losses decrease sharply and consistently, indicating strong learning capabilities and effective generalization without significant signs of overfitting. In contrast, Figure 4 (mbart50) and Figure 6 (mt5) display effective initial learning but suggest potential overfitting in later stages, as evidenced by the validation losses that either flatten or increase while training losses continue to decrease. These observations underscore the need for precise model tuning and the adoption of strategies such as regularization and data augmentation to enhance the models' performance and ensure their ability to generalize to new, unseen data.

## 5    Conclusion

We created a specialized dataset of biology textbooks for 11th and 12th grade students to improve the understanding of advanced linguistic models such as BERT and MT5 for educational content. The study focuses on training NLP models, specifically AraBERT, mBart50, araT5, and MT5, using this specialized suite to enhance their ability to understand and produce educational content efficiently. Comprehensive evaluation criteria were adopted to evaluate the performance of these models, including manual evaluation by experts in education and textbook writing, and automated evaluation metrics such as ROUGE.

Future work in NLP looks to push the boundaries toward achieving a deeper and more accurate understanding of complex scientific texts, focusing on improving the ability of machine models to summarize these texts efficiently. This includes developing new technologies to overcome the unique linguistic challenges presented by the Arabic language and expanding the scope of training databases to cover a more comprehensive range of scientific topics and concepts. This direction also includes improving the methods used in evaluations, whether automated or manual, to ensure the production of high-quality and effective summaries. In addition, advanced optimization techniques such as LoRA (Low-Rank Adaptation) are being explored to enhance the performance of trained models, promising significant advances in how scientific texts in Arabic are understood and processed in the future.

**ACKNOWLEDGMENT**

We are deeply grateful to Mr. Ayman Raddad - 16 years' experience in teaching Biology for 11th and 12th grades - and the Palestinian Ministry of Education for providing all the summaries and evaluations as experts to conduct this study.